\documentclass[preprint]{ccn}
\usepackage{amsmath}
\usepackage{booktabs}
\usepackage[table]{xcolor}
\usepackage{tabularx}
\usepackage{array}
\usepackage{placeins}
\usepackage{caption}
\usepackage{dblfloatfix}
\definecolor{darkgreen}{rgb}{0.0,0.5,0.0}

\makeatletter
\renewcommand\subsubsection{\@startsection{subsubsection}{3}{\z@}%
  {1.0ex}%
  {0.6ex}%
  {\normalfont\normalsize\bfseries}}%
\makeatother

\title{Appearance-free Action Recognition: Zero-shot Generalization in Humans and a Two-Pathway Model}

\author{Prerana Kumar\affmark{1,2}, Martin A. Giese\affmark{1}}
\affiliation{1}{Hertie Institute, University of Tübingen}
\affiliation{2}{IMPRS-IS}
\emails{prerana.kumar@uni-tuebingen.de}

\makeatletter
\renewenvironment{abstract}
{
  {\centering\bfseries\Large Abstract\par}
  \vspace{0.25em}
  \list{}{\leftmargin=0.02\columnwidth \rightmargin=0.02\columnwidth}
  \item\relax
  \normalsize
}
{
  \endlist
  \vspace{0em}
}
\makeatother
\begin{document}

\maketitle

\begin{abstract}
Action recognition is a fundamental ability for social species. However, the computations underlying robust action recognition are not well understood.  Classical psychophysical studies using simplified or artificial stimuli have shown that humans can perceive body motion even under severe degradation of relevant shape cues. Recent work using real-world action videos and their appearance-free counterparts (that preserve motion but lack static shape and texture cues) included explicit training of humans and models on the appearance-free videos. These studies do not address the question of whether humans and vision models generalize in a zero-shot manner to appearance-free transformations of real-world action videos. To address this gap, we measured this generalization in a laboratory-based psychophysics experiment. 22 participants were trained to recognize five action categories using only naturalistic videos (UCF5 dataset), and tested zero-shot on two types of appearance-free transformations: (i) dense-noise motion videos from an existing appearance-free dataset (AFD5) and (ii) random-dot appearance-free videos generated by us. We find that participants recognize actions in both types of appearance-free videos well above chance levels, albeit with reduced accuracy compared to the naturalistic videos. To model this behavior, we developed a two-pathway 3D CNN-based model combining an RGB (form) stream and an optical flow (motion) stream, which includes a coherence-gating mechanism inspired by Gestalt common-fate grouping. The model was trained on UCF5 videos and evaluated zero-shot on the appearance-free datasets. It performs well above chance on both appearance-free datasets and outperforms contemporary video classification models, narrowing the performance gap between models and humans. Ablation experiments reveal that the motion pathway is critical for generalization to the appearance-free videos, while the form pathway improves overall performance on naturalistic videos. Our findings highlight the importance of motion-based representations for robust generalization to appearance-free videos, and support the use of multi-stream architectures to model video-based action recognition.
\end{abstract}

\begin{figure*}[t]
  \centering
  \includegraphics[width=\textwidth]{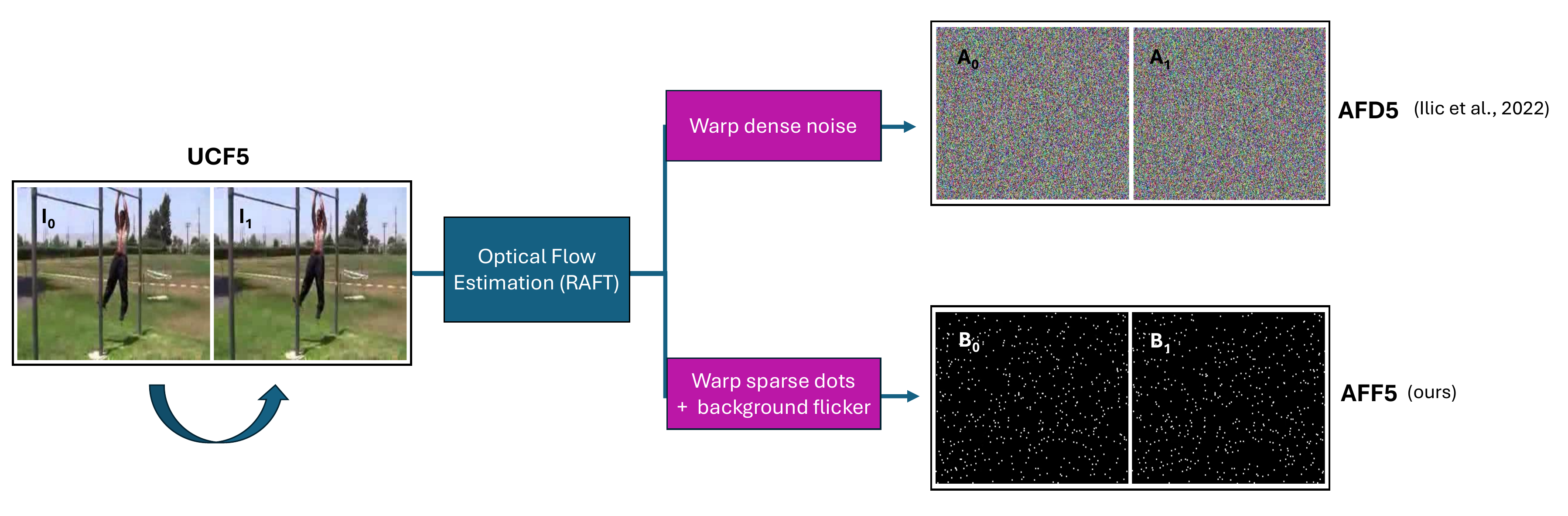}
  \caption{\textbf{Appearance-free stimulus generation.} AFD5 videos \citep{Ilic2022} are generated by warping dense noise through time using RAFT optical flow \citep{TeedDeng2020RAFT} from the source RGB video. AFF5 videos (this work) are generated by warping sparse dots with the same RAFT flow and introducing finite-lifetime random dots in the background}
  \label{fig:dataset_generation}
\end{figure*}

\begin{figure*}[t]
  \centering
  \includegraphics[width=\textwidth]{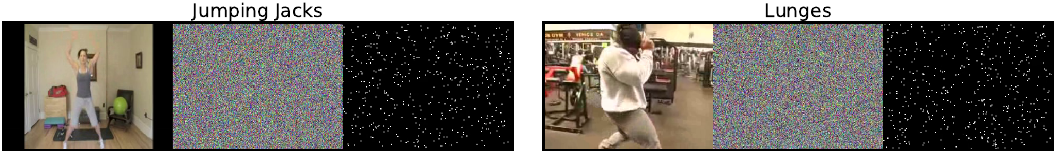}
  \caption{\textbf{Example stimuli.} Representative single frames from two action videos across UCF5 (RGB), AFD5 (dense-noise appearance-free), and AFF5 (sparse random-dot appearance-free) videos. Single frames in AFD5/AFF5 videos contain minimal static appearance cues}
  \label{fig:stimuli_examples}
\end{figure*}

\vspace*{-0.4em}
\section{Introduction}

Humans excel at recognizing actions in videos despite large variations in appearance, viewpoint, and background context. Yet the detailed computational mechanisms underlying robust action recognition are not well understood. 

Motion provides a particularly powerful cue for action recognition and object segmentation from video stimuli. In human vision, motion segmentation is often linked to the Gestalt principle of common fate \citep{Wertheimer1912,Wertheimer2012}, which posits that elements that move coherently are perceptually grouped as belonging together. Previous work shows that action perception may not be purely motion-based: even highly reduced displays can contain sparse posture or configural information that becomes informative through temporal integration. Multiple studies propose that biological-motion recognition can be achieved using motion features, form-over-time, or their integration \citep{WebbAggarwal1982,ChenLee1992,CasileGiese2005,BeintemaLappe2002,LangeLappe2006,GiesePoggio2003}. 
To isolate these specific mechanisms, prior studies often utilized stimuli lacking certain texture or appearance cues. Classical work in psychophysics demonstrates that humans can recognize biological motion from videos even in the absence of recognizable visual shape and texture cues. For example, Johansson’s seminal work \citep{Johansson1973} using point-light displays yields vivid percepts of biological motion from a small set of moving dots. Later work showed that humans could recognize “formless” dot-field structure-from-motion stimuli \citep{SingerSheinberg2008}, constructed such that individual frames resemble random noise while coherent objects or animated figures emerge only through the dynamic integration across subsequent frames. \citet{Tangemann2024} showed that both humans and a neuroscience-inspired motion-energy model can perform random-dot motion segmentation with zero-shot generalization to novel synthetic stimuli preserving the object motion from the original videos. Recent findings from neuroscience \citep{RobertUngerleiderVaziriPashkam2023} demonstrated that motion cues in the form of kinematograms can elicit widespread and robust category responses, even in ventral regions of the visual cortex that have traditionally been associated mainly with image-defined form processing. 

However, much of the existing evidence for recognition of biological motion from visually degraded stimuli relies on tightly controlled or simplified, artificially generated stimuli. How strongly these conclusions generalize to videos including the background clutter, camera motion, and variability characteristic of real-world action datasets remains to be investigated. A major step toward bridging this gap is the recently introduced Appearance Free Dataset (AFD) \citep{Ilic2022}, which is comprised of “appearance-free” action videos generated by estimating optical flow between consecutive frames of real-world RGB videos and using it to warp a random noise frame image through time, ensuring that single frames contain no static appearance cues while inter-frame motion is preserved. Importantly, prior human psychophysical experiments with these appearance-free action videos evaluated performance after participants were explicitly trained to categorize the appearance-free videos (in addition to RGB video training), and the proposed two-stream model was likewise trained on the appearance-free videos. Recent work comparing macaque IT responses and contemporary video ANNs has suggested that current video models remain limited in their ability to generalize across appearance-free video transformations preserving object motion, while primate ventral-stream activity demonstrates this generalization \citep{DunnhoferMicheloniKar2026}. It remains unclear how well humans and contemporary video recognition models generalize in a zero-shot manner from naturalistic RGB videos of human actions to their appearance free counterparts. In this study, we aim to investigate and model human action perception using apperance-free videos of naturalistic actions.

To this end, we measure zero-shot generalization to appearance-free actions exhibiting real-world variability and probe robustness when stable-scene segmentation cues are disrupted. Specifically, we make the following key contributions: 
\begin{enumerate}
  \item We measure human zero-shot generalization to existing appearance-free videos (AFD5, a 5-action subset of the AFD dataset \citep{Ilic2022}) in an action classification task, following training on only RGB videos (UCF5 dataset, the corresponding subset of the UCF101 human action dataset \citep{Soomro2012UCF101}). To test generalization when the remaining form cues in AFD5, which result from motion segmentation, are disrupted, we developed a third random-dot appearance-free data set, that we refer to as AFF5. These videos contain randomly appearing dots with finite lifetimes in the background. We show that human participants perform well above chance on the recognition task, although with lower accuracy on the appearance-free stimuli than on the original RGB videos.
  \item We develop a two-stream 3D CNN model, CG2-X3D, that integrates an RGB (“form”) pathway with a motion pathway, which includes an optical flow estimator and an additional coherence-gating mechanism that is motivated by the Gestalt principle of common fate. The model generalizes well above chance to both the appearance-free datasets without any prior training on such videos.
  \item Finally, we benchmark our model against contemporary video recognition architectures and show that the proposed two-stream model yields substantially improved zero-shot transfer to appearance-free videos, significantly reducing the gap to human performance.
\end{enumerate}

\section{Methods}

\begin{figure*}[t]
  \centering
  \vspace{0.5\baselineskip}
  \includegraphics[
    width=\textwidth,
    height=0.22\textheight,
    trim={1.2cm 0.6cm 1.2cm 1cm},
    clip
  ]{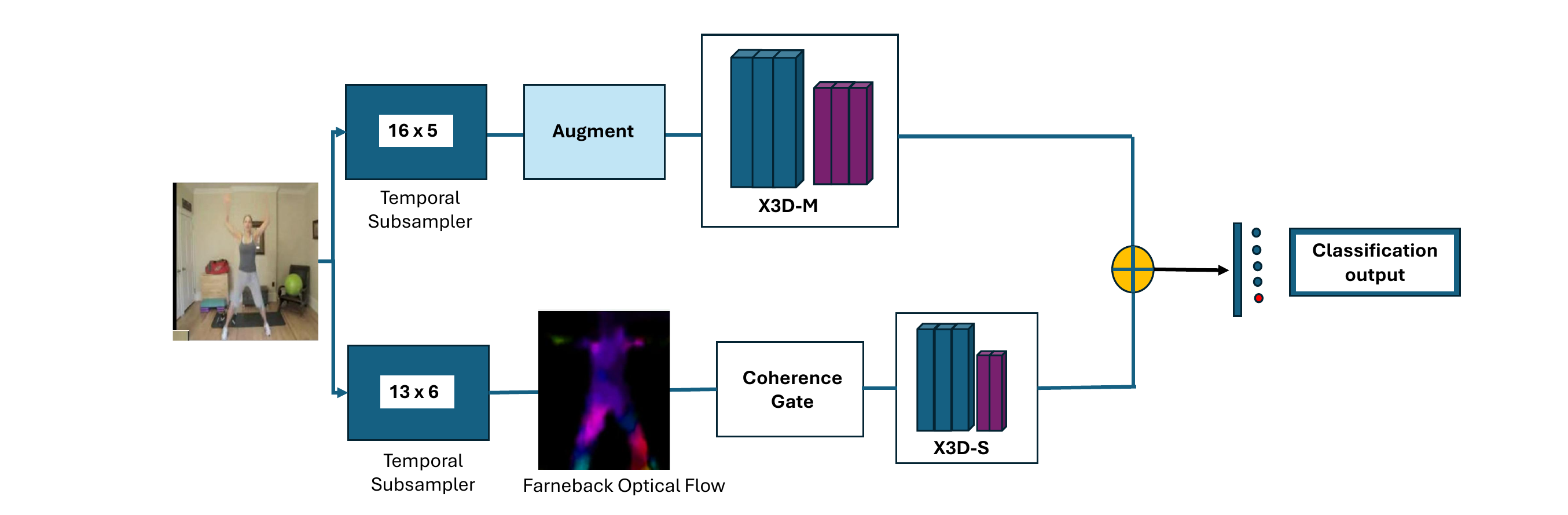}
  \caption{\textbf{Model schematic (CG2-X3D).} Two-stream 3D CNN-based architecture with an RGB (form) stream and an explicit motion stream with fusion of streams for classification. The motion stream includes coherence-gated optical flow representations.}
  \label{fig:model_schematic}
\end{figure*}

\subsection{Stimuli}

UCF5 has been used in prior work on appearance-free action recognition \citep{Ilic2022}, and comprises of five action categories: Pushups, Pullups, Swings, Lunges, and Jumping Jacks. For model training and evaluation we used a data split with 409 training videos and 174 test videos across all categories. Model training used the full UCF5 training split, whereas the human psychophysics experiment used a smaller labeled training set (4 training videos per class) and test set (15 videos per class).

AFD5 consists of videos from the Appearance Free Dataset corresponding to UCF5. These videos were generated by estimating optical flow between adjacent RGB frames using RAFT \citep{TeedDeng2020RAFT} and warping a random noise image through time, yielding dense-noise videos whose inter-frame motion matches the source video while static appearance cues are removed (Fig.~\ref{fig:dataset_generation}). In those videos in which there is no camera motion or background object/human motion, the background pixels remain static. This facilitates motion segmentation of the moving body against the static background.

AFF5 was generated from RAFT optical flow tensors corresponding to the same underlying UCF5 videos (Fig.~\ref{fig:dataset_generation}). Each AFF5 video is a sparse random-dot stimulus created by warping a frame containing sparse dots using the RAFT flow field. During the video, finite-lifetime dots appear in random locations in the background. We initialize 500 dots uniformly across the frame on a black background. At each time step, dot velocities are obtained by bilinear sampling of the flow at each dot location and dot positions are updated accordingly. Dots appear at new, random locations when their lifetime expires (8 frames) or when they move out of the image boundaries. The finite lifetime background dots disrupt the motion segmentation present in the AFD5 videos, while preserving motion signals corresponding to the actor performing the action.
Examples of static frames from the three stimulus sets are shown in Fig.~\ref{fig:stimuli_examples}.

\subsection{Human psychophysics experiment}

We collected classification accuracy and reaction time data from 22 participants (13 female, 9 male) in a controlled psychophysics lab setting. The experiment location, computer/display, lighting and seating distance from the display (0.6 meters) were kept constant for all participants. Participants included both trained vision scientists and naive subjects. The experiment was implemented in PsychoPy \citep{Peirce2007PsychoPy} as a 5-way forced-choice classification task. Participants were trained with labeled UCF5 videos (4 per class) to familiarize them with the classes and each training video auto-looped twice before advancing. Participants then completed three evaluation blocks (without feedback on their answers) using 75 selected clips from each of the three datasets: UCF5 (RGB), AFD5, and AFF5. AFD5 and AFF5 were transformations of the same underlying RGB test videos. The UCF5 evaluation block served the purpose of allowing participants to get familiar with the task, and as a test that participants had a clear understanding of the various action classes before performing the generalization test. Videos used in the experiment were pre-screened and selected to avoid ceiling or floor effects, and to balance difficulty across classes to the degree possible. Importantly, participants received no training on appearance-free stimuli prior to the appearance-free blocks. All participants completed the training block first, followed by the UCF5 test block. The order of the two appearance-free blocks was counterbalanced (half the subjects viewed AFD5 videos before AFF5 and the other half viewed the videos in the reverse order). The 75 videos per dataset were arranged into five sub-blocks of 15 videos, each balanced across classes (3 videos per class), and video ordering was varied across participants, subject to this constraint. The test videos looped until response, transitioning after 30s to the response screen if no response had been provided until then. A two-press paradigm (spacebar to indicate recognition followed by category choice from the keyboard) was implemented to record accurate reaction times (RTs), followed by the classification response. Participants were instructed to press the space-bar as soon as they recognized the action category, but to take their time selecting the key corresponding to the recognized class to minimize misclassifications due to key press errors. 

\subsection{Two-Stream Model: CG2-X3D}
Our proposed model, CG2-X3D (Fig.~\ref{fig:model_schematic}), is a two-stream 3D CNN with an RGB stream (X3D-M) and a motion stream (X3D-S) that also includes an optical flow estimator and a coherence gate. The X3D \citep{Feichtenhofer2020X3D} network in the motion stream receives input in the form of explicit optical-flow representation rendered as a 3-channel HSV visualization. To suppress background random-dot flickering in motion representations, we apply coherence-based attenuation (“coherence gate”) to the output of the optic flow estimation stage, downweighting locally incoherent orientation fields and low-magnitude motion. Each stream of the model produces a spatiotemporally pooled feature vector, which is projected to a 512-dimensional embedding via a linear reduction layer, L\textsubscript{2}-normalized, concatenated across streams, and mapped to class logits via a linear classifier head. For the main configuration, the RGB stream used 16 frames (sample rate = 5, input size = $224 \times 224$), and the motion stream used 13 frames (sample rate = 6, input size = $182 \times 182$).

\subsection{Benchmarking}
We compared against single-stream and two-stream contemporary video recognition models (e.g., I3D \citep{CarreiraZisserman2017}, SlowFast \citep{Feichtenhofer2019SlowFast}, MViTv2 \citep{Li2021MViTv2}, TimeSformer \citep{Bertasius2021TimeSformer}) and additional baselines used in ablations.  We also compared our model to a version in which the coherence-gated optical flow front end was replaced with a motion-energy front end based on Tangemann et al.’s implementation of the Simoncelli-Heeger model \citep{SimoncelliHeeger1998,Tangemann2024}. Additionally, we evaluated a frozen, pretrained backbone of a recent, state-of-the-art image model from computer vision, DINOv3 \citep{Simeoni2025DINOv3}. We froze the DINOv3 backbone and trained a temporal head module that aggregates frame-level embeddings over time (mean pooling) followed by a linear classifier to predict the action label. For each video, we loaded the RGB input and (when applicable) its corresponding optical flow encoding, using the same UCF split files. For each stream, we computed the target clip duration from the video’s frame-rate metadata and the effective number of sampled frames $(\mathrm{num\_frames} \times \mathrm{sample\_rate})$, then sampled a temporally aligned window from both streams. For the frozen DINOv3 baseline, RGB videos were converted back to pixel space and then normalized using the DINO image processor statistics required by the pretrained backbone. Training-time RGB videos used probabilistic grayscale conversion ($p = 0.2$) and a small temporal jitter ($\pm 1$ frame after subsampling). During two-stream training, we applied a stochastic horizontal flipping augmentation with probability 0.5 synchronously to both the RGB and motion streams. For HSV-encoded optical flow videos, we additionally remapped the hue after flipping to maintain correct motion-direction encoding under horizontal reflection. Evaluation used deterministic transforms (no jitter, grayscale or video flip augmentations), with centered video sampling.

\subsection{Motion-stream optical flow and coherence gate }
\label{sec:flow_gate}

We computed dense optical flow between consecutive frames using the Farnebäck algorithm \citep{Farneback2003} and rendered the flow in HSV format: hue encodes flow direction, saturation is fixed (255), and value encodes motion magnitude after normalization. For each flow field, we first computed the per-pixel optical flow magnitude. We then normalized each frame by using a frame-dependent reference value $d_t$, defined as a weighted average of the 95th-percentile optical flow magnitude estimated on the UCF5 training set, and the current frame’s 95th-percentile flow magnitude. To suppress extremely small motions, we first subtracted a small offset from the magnitude before normalization, where the offset was set to 3\% of the per-frame normalization scale. The resulting magnitudes were then linearly mapped to $[0,1]$ and bounded to this interval, producing the pre-gate value map.

\paragraph{Coherence gate}
 Let $\mathbf{f}_t(\mathbf{x})=(u_t(\mathbf{x}),v_t(\mathbf{x}))$ denote the Farnebäck flow and $m_t(\mathbf{x})=\|\mathbf{f}_t(\mathbf{x})\|_2$ its magnitude. We computed a local orientation coherence from the unit flow vectors
\begin{equation}
\mathbf{n}_t(\mathbf{x}) = \frac{\mathbf{f}_t(\mathbf{x})}{m_t(\mathbf{x}) + \varepsilon}.
\end{equation}
where $\varepsilon$ is a small positive constant to prevent division by zero. 
Let $\mathcal{N}_k(\mathbf{x})$ denote a $k \times k$ neighborhood of size $|\mathcal{N}_k|$. We computed the mean unit-vector in this window ($k=9$) via a box filter on the unit-vector components,
\begin{equation}
\bar{\mathbf{n}}_t(\mathbf{x}) = \frac{1}{|\mathcal{N}_k(\mathbf{x})|}
\sum_{\mathbf{y}\in\mathcal{N}_k(\mathbf{x})}\mathbf{n}_t(\mathbf{y}),
\end{equation}
where coherence is defined as the magnitude of the resultant vector,
\begin{equation}
c_t(\mathbf{x}) = \big\|\bar{\mathbf{n}}_t(\mathbf{x})\big\|_2 \in [0,1].
\end{equation}

We converted the raw coherence score into a multiplicative gating weight by applying a soft threshold with threshold value $\tau=0.30$: values below $\tau$ were set to zero, and values above $\tau$ were linearly rescaled to the interval $[0,1]$. In parallel, we computed a magnitude-dependent weight from the flow magnitude normalized by the same per-frame reference value $d_t$ used for the HSV value-channel normalization: the normalized magnitude was bounded to $[0,1]$ and hard-thresholded so motions below $r_{\min}=0.02$ (as a fraction of $d_t$) were set to zero. The instantaneous coherence-gate mask is defined as the product of the magnitude term and the thresholded coherence term, with the latter raised to a coherence-weighting exponent $\beta$, such that larger $\beta$ values produced stronger suppression of partially coherent motion. We then temporally smoothed this mask across frames using an exponential moving average (coefficient $\lambda=0.80$, initialized with the first frame of each video),
\[
g_t(x) = \lambda\, g_{t-1}(x) + (1-\lambda)\, g_t^{\mathrm{inst}}(x).
\] where the instantaneous mask is denoted by $g_t^{\mathrm{inst}}(x)$ and its temporally smoothed version by $g_t(x)$.
Finally, we attenuated the HSV value channel by multiplying the pre-gate value map $V^{(0)}$ by this smoothed mask (with clipping to $[0,1]$) and converted the HSV representations back to standard BGR format as input to the downstream X3D network.
To keep the modeling strictly zero-shot, we did not tune the hyperparameters of the coherence gate on the appearance-free datasets.

\subsection{Optimization, training, and evaluation}

CNN baselines and the two-stream model were trained with Adam (lr $= 3\times 10^{-4}$) using label-smoothed cross entropy ($\epsilon=0.1$) and gradient clipping (max norm 5.0). The transformer benchmarks (MViTv2 and TimeSformer) needed to be fine-tuned with an AdamW \citep{LoshchilovHutter2019AdamW} schedule, as these architectures are more sensitive to optimization under short schedules. All other data and augmentation settings were held constant. The frozen DINOv3 probe used the same AdamW configuration, applied to the probe head parameters only. All models were finetuned for 10 epochs on the UCF5 RGB training set and evaluated on the test set and its corresponding AFD5/AFF5 transformations. We ran five random seeds (42–46) and report the mean $\pm$ SD across seeds. We report results from a fixed checkpoint (the 10th epoch) for all seeds and models. The models were initialized with Kinetics-pretrained PyTorchVideo weights. After training, each model was evaluated on: (i) UCF5 test set, and (ii) the AFD5 and AFF5 versions of these test exemplars. Performance is reported as top-1 accuracy (chance = 20\%). We additionally compute a “Transfer Score” defined as the mean of zero-shot AFD5 and AFF5 accuracy.

\section{Results}

Our goal was to test human performance on appearance-free action videos after being trained only on naturalistic RGB videos, and to develop a model that can reproduce or narrow the gap to human performance. Below, we first report human performance across the three conditions that we tested, then compare a range of models trained only on RGB, and finally use ablations to isolate the contributions of the form and motion pathways and the coherence gate.

\subsubsection{Humans generalize to real-world appearance-free videos without any appearance-free training.} 
\begin{figure*}[t]
  \centering
  \begin{minipage}[t]{0.49\textwidth}
    \centering
    \includegraphics[width=\linewidth,height=0.30\textheight,keepaspectratio]{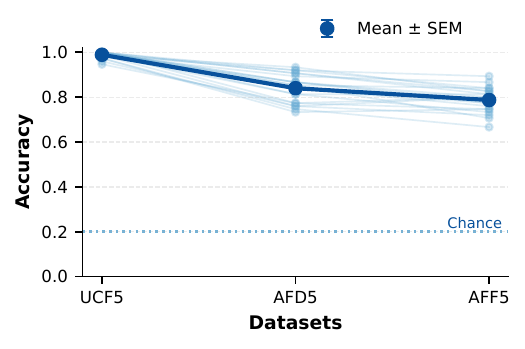}
    \caption{\textbf{Human accuracy across conditions.} Each light blue point/line shows one participant’s accuracy across UCF5, AFD5, and AFF5. The dark blue points/line indicate the group mean $\pm$ SEM across participants ($n=22$). The dotted horizontal line indicates chance accuracy (20\%).}
    \label{fig:human_accuracy}
  \end{minipage}
  \hfill
  \begin{minipage}[t]{0.49\textwidth}
  \centering
  \includegraphics[width=\linewidth,height=0.28\textheight,keepaspectratio]{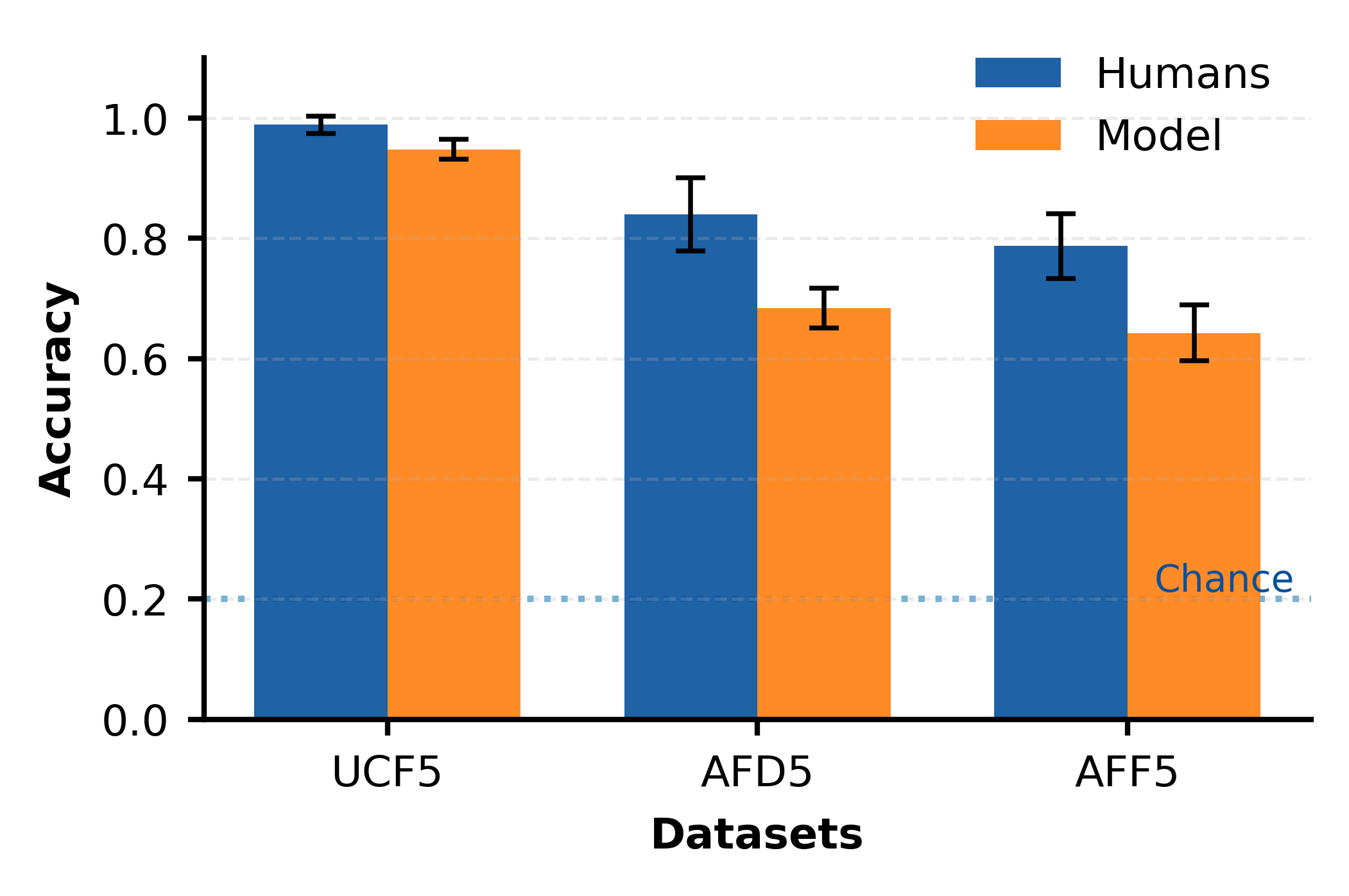}
  \caption{\textbf{Humans vs model.} Mean accuracy on UCF5, AFD5, and AFF5 for human observers ($n=22$) and the proposed model (mean across 5 random seeds). Human error bars indicate $\pm$SD across participants; model error bars indicate $\pm$SD across seeds. Chance performance is 20\%.}
  \label{fig:humans_vs_model}
\end{minipage}
\end{figure*}

~\\
Humans generalized strongly to both types of real-world appearance-free videos without any appearance-free training, but performance decreased reliably relative to naturalistic RGB videos (Fig.~\ref{fig:human_accuracy}; Fig.~\ref{fig:humans_vs_model}). Mean accuracy across participants was $0.9891 \pm 0.0146$ on UCF5, $0.8400 \pm 0.0613$ on AFD5, and $0.7872 \pm 0.0539$ on AFF5. A repeated-measures ANOVA revealed a strong effect of stimulus condition on accuracy (UCF5, AFD5, AFF5), $F(2,42)=188.39$, $p<10^{-10}$, partial $\eta_p^2=0.900$. This effect was confirmed by a nonparametric Friedman test, $\chi^2(2)=38.55$, $p=4.25\times 10^{-9}$. Group means followed the ordering $\text{UCF5} > \text{AFD5} > \text{AFF5}$. Importantly, accuracy in all conditions remained well above chance (20\%). This ordering held for most participants, although a small subset (3 out of 22) showed the reverse pattern between the two appearance-free conditions ($\text{AFF5} > \text{AFD5}$), and 1 participant achieved equal accuracy on both appearance-free datasets. To account for sequence effects or incidental learning from exposure to appearance-free stimuli, we counterbalanced block order and divided participants into two groups (AFD5-first vs.\ AFF5-first, $N=11$ in each group). The mean performance on AFD5 was higher than AFF5 in both groups (AFD5-first: $0.8533 \pm 0.0540$ accuracy on AFD5 vs.\ $0.8036 \pm 0.0481$ on AFF5, paired $t(10)=4.12$, $p=0.0021$; AFF5-first: $0.8267 \pm 0.0677$ on AFD5 vs.\ $0.7709 \pm 0.0565$ on AFF5, paired $t(10)=3.41$, $p=0.0067$), and the AFD5 -- AFF5 difference did not differ by block order (Welch $t=-0.298$, $p=0.769$), providing no evidence for a systematic ordering effect. These results show that humans exhibit robust appearance-free action recognition, while revealing a consistent group-level cost of disrupting stable scene priors in the videos.

\subsubsection{Benchmark video models show large drops upon removal of appearance cues despite strong RGB performance. }
 \citet{Ilic2022} showed that appearance-free action recognition remains challenging for contemporary video models, even when models are trained directly on the appearance-free videos. Consistent with this, we found that zero-shot transfer from RGB videos to appearance-free action videos is not trivial for contemporary architectures trained only on naturalistic videos. All benchmark video models models were initialized from Kinetics-pretrained weights, fine-tuned on UCF5 RGB training videos, and then evaluated on the held-out UCF5 test set and its appearance-free counterparts (AFD5 and AFF5). Because DINOv3 is a recent state-of-the-art model with substantial pretraining, we included it as a baseline using a frozen-feature protocol. We evaluated a diverse set of baselines spanning single-stream and two-stream architectures, including models with explicit optical flow inputs (e.g., I3D and E2SX3D) and a motion-energy model. For the motion-energy baseline, we replaced our flow-and-gating front end with a motion-energy representation while keeping the downstream two-stream classification pipeline fixed. Across these comparisons, many models achieved excellent performance on UCF5, yet showed large drops on AFD5 and AFF5 (Table 1.). In contrast, our coherence-gated two-stream model achieved the strongest appearance-free transfer, indicating that high RGB accuracy - even in architectures with explicit optical flow inputs - is not sufficient for robust appearance-free generalization.

\subsubsection{CG2-X3D shows the highest generalization performance to appearance-free videos and significantly narrows the human-model gap. }
When trained only on UCF5 RGB videos, CG2-X3D generalizes well to both appearance-free datasets and clearly outperforms all the tested baselines in appearance-free action recognition, including state-of-the-art video recognition models and other two-stream models including explicit optical flow. It also replicates the performance ordering observed in humans (UCF5 $>$ AFD5 $>$ AFF5). Despite these gains, CG2-X3D narrows but does not fully close the gap to human performance (Fig.~\ref{fig:humans_vs_model}).

We tested whether the coherence-gating mechanism contributes beyond the presence of an explicit motion stream by comparing CG2-X3D to an otherwise matched version of the model without coherence gating (Fig.~\ref{fig:ablations}; Table~\ref{tab:benchmarks}). Removal of the gate produced a clear drop in appearance-free transfer, with the largest degradation in the random-dot condition (AFF5). This aligns with the intended function of the gate: downweighting locally incoherent, transient motion while preserving coherent motion patterns, typically associated with the actor. Thus, a simple mechanism based on local coherence provided a significant boost to the overall robustness of the model to the removal of appearance cues.

\subsubsection{Ablations reveal complementary roles of RGB and motion pathways. }
\begin{figure}[t]
  \centering
  \includegraphics[width=\columnwidth]{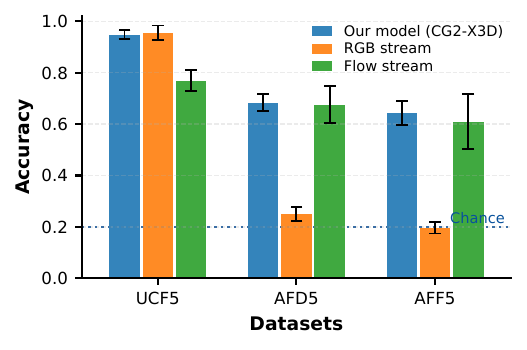}
  \caption{\textbf{Ablations.} Mean accuracy across five seeds for RGB-only model, flow-only model, and the two-stream model across UCF5, AFD5, and AFF5. Error bars indicate ±1 SD across seeds.}
  \label{fig:ablations}
\end{figure}

To isolate which model components support appearance-free transfer, we evaluated ablated variants of CG2-X3D (Fig.~\ref{fig:ablations}; Table~\ref{tab:benchmarks}). An RGB-only model achieved high accuracy on naturalistic UCF5 videos but collapsed to near-chance performance on both appearance-free conditions, consistent with the removal of static shape/texture cues in AFD5 and AFF5. In contrast, a motion-only model driven by explicit optical flow preserved the strong generalization to AFD5 and AFF5 observed in the two-stream model, but did not perform as well as the two-stream model on UCF5, indicating that motion representations strongly support appearance-free action recognition, yet are incomplete for optimal performance on naturalistic UCF5 videos. The full two-stream model combined these complementary strengths, maintaining high UCF5 performance while also retaining robust appearance-free accuracy. Together, these findings align with evidence from vision science that motion and form pathways interact during perception, and support the use of multi-stream architectures as models of robust visual processing.

\begin{table*}[!t]
\centering
\caption{Action recognition accuracy (mean $\pm$ SD across 5 seeds) on UCF5 (RGB) and the corresponding appearance-free test sets (AFD5, AFF5).
Transfer Score is $(\mathrm{AFD5}+\mathrm{AFF5})/2$ computed from the reported means .}
\label{tab:benchmarks}
\footnotesize
\setlength{\tabcolsep}{12pt}
\renewcommand{\arraystretch}{1.25}

\rowcolors{2}{gray!4}{white}

\begin{tabularx}{\textwidth}{@{}>{\raggedright\arraybackslash}X r r r c@{}}
\toprule
\rowcolor{white}
\textbf{Model} &
\multicolumn{1}{c}{\textbf{UCF5}} &
\multicolumn{1}{c}{\textbf{AFD5}} &
\multicolumn{1}{c}{\textbf{AFF5}} &
\multicolumn{1}{c}{\textbf{Transfer Score}} \\
\midrule
CG2-X3D (two-stream + coherence gate) & 0.9483 $\pm$ 0.0163 & \textbf{0.6839 $\pm$ 0.0330} & \textbf{0.6425 $\pm$ 0.0464} & \textbf{0.6632} \\
CG2-X3D (RGB stream-only)             & 0.9552 $\pm$ 0.0286 & 0.2494 $\pm$ 0.0268 & 0.1965 $\pm$ 0.0220 & 0.2230 \\
CG2-X3D (Flow stream-only)            & 0.7689 $\pm$ 0.0411 & 0.6758 $\pm$ 0.0719 & 0.6092 $\pm$ 0.1072 & 0.6425 \\
CG2-X3D (no coherence gate)           & 0.9644 $\pm$ 0.0286 & 0.6069 $\pm$ 0.0507 & 0.3345 $\pm$ 0.0724 & 0.4707 \\
\midrule
TimeSformer                           & 0.9391 $\pm$ 0.0051 & 0.2402 $\pm$ 0.0228 & 0.2046 $\pm$ 0.0180 & 0.2224 \\
MViTv2                                & \textbf{1.0000 $\pm$ 0.0000} & 0.2678 $\pm$ 0.0515 & 0.2092 $\pm$ 0.0246 & 0.2385 \\
SlowFast                              & 0.9092 $\pm$ 0.0245 & 0.3046 $\pm$ 0.0370 & 0.2184 $\pm$ 0.0129 & 0.2615 \\
DINOv3 (mean, frozen probe)           & 0.8069 $\pm$ 0.0155 & 0.2414 $\pm$ 0.0203 & 0.2034 $\pm$ 0.0406 & 0.2224 \\
I3D                                   & 0.9471 $\pm$ 0.0249 & 0.3368 $\pm$ 0.0221 & 0.2299 $\pm$ 0.0548 & 0.2834 \\
E2S-X3D                               & 0.9966 $\pm$ 0.0051 & 0.5241 $\pm$ 0.0889 & 0.4207 $\pm$ 0.0596 & 0.4724 \\
Motion energy model (two-stream)      & 0.9460 $\pm$ 0.0175 & 0.6678 $\pm$ 0.0228 & 0.2448 $\pm$ 0.0302 & 0.4563 \\
\bottomrule
\end{tabularx}

\rowcolors{2}{}{} 
\end{table*}

\section{Discussion}

 Our behavioral experiment shows that humans can categorize real-world, highly variable actions from appearance-free videos well above chance without any training on appearance-free stimuli. This extends classical findings from simplified motion displays to naturalistic, highly variable stimulus classes. Prior work in this area evaluated appearance-free recognition using AFD5 stimuli, which do not prevent form-from-motion segmentation as they include static backgrounds (in examples without camera or background object motion). Thus, we developed a new random dot appearance-free stimulus set using the UCF5 videos, that greatly reduces such segmentation. This manipulation targets a key ambiguity in AFD5 videos: although static appearance cues are removed, stable backgrounds can still allow segmentation-like form-from-motion structure to emerge over time. Importantly, performance on AFF5 remained well above chance, indicating that while such cues are helpful, observers can rely on the actor’s motion to support robust action recognition even when stable-scene regularities are disrupted. Nevertheless, performance on AFF5, while strong, is slightly lower than on AFD5. This suggests that motion cues support substantial generalization, but disrupting segmentation-from-motion cues still has a measurable impact on accuracy in real-world video-classification tasks. 

Deep neural networks have become central computational models in both cognitive science and computer vision, and are used frequently to account for human perception and behavior. In vision, feedforward convolutional and recurrent architectures optimized for recognition tasks can capture a range of behavioral signatures and, in many settings, predict neural responses across the visual hierarchy. At the same time, strong performance within a training distribution does not guarantee human-like generalization to out-of-domain inputs. A key challenge, therefore, is to identify stimulus conditions under which model robustness diverges from human robustness, and to use those divergences to motivate more mechanistic model components. Standard video recognition models trained only on RGB videos did not reproduce the zero-shot generalization ability of humans: despite strong performance on naturalistic videos, most architectures generalized poorly to AFD5 and AFF5. 

To address this limitation, we developed a coherence-gated two-stream model (CG2-X3D), which achieved substantially higher zero-shot accuracy on the appearance-free videos and narrowed the gap to human performance. Ablations indicate a complementary division of labor between streams: the motion pathway is critical for appearance-free generalization, whereas the RGB pathway supports high in-domain performance on naturalistic videos. The coherence gate provides an additional boost, most notably when segmentation-like cues are disrupted, by increasing the influence of locally coherent motion. There is strong evidence that the human dorsal pathway contains neural structures that extract locally coherent motion \citep{NewsomePare1988}.

\subsubsection{Limitations. } Firstly, we focus on a small action set (five classes), chosen to enable controlled lab psychophysics and tightly matched stimulus manipulations. It is known that humans can hold five action classes in working memory for a prolonged duration (\cite{Miller1956}), making five class datasets particularly suitable for such an experiment. The extent to which the same human–model robustness patterns hold at larger scale remains to be tested.   Secondly, while it outperforms other tested models and achieves accuracy levels well above chance on the appearance free datasets, a definite generalization gap to human level performance persists. Although participants received no labeled training on appearance-free stimuli, they may still exhibit unsupervised learning effects due to exposure during the testing block, whereas models are evaluated without any test-time adaptation and do not receive this advantage. Finally, unlike humans, due to the large number of parameters in deep neural network models, the models require significantly more training data than humans.

Our results suggest concrete directions for future work. On the modeling side, an immediate test is to evaluate zero-shot generalization using coherence gating across larger action sets. On the behavioral side, extending our psychophysical study to more actions and to fully novel appearance-free exemplars (rather than transformations of previously seen RGB test videos) would provide an even more challenging out-of-domain transfer test. Adding additional neuroscience-inspired features to existing models and using such models to analyze human and model stimulus difficulty rankings could provide additional insights into mechanisms used in action recognition. Finally, the framework generates neural-predictions: appearance-free recognition should place greater demands on motion-selective/dorsal processing, and the finite-lifetime random dot manipulation should increase reliance on computations that emphasize coherent motion structure. Another promising future direction in this zero-shot generalization task is to link model uncertainty to human reaction time and confidence, providing an additional behavioral constraint on the decision process in response to appearance free stimuli.

\section{Acknowledgments}

PK and MG were supported by the ERC (SyG 856495). MG was additionally supported by HFSP (RGP0036/2016) and BMBF (FKZ 01GQ1704). We thank the International Max Planck Research School for Intelligent Systems (IMPRS-IS) for supporting Prerana Kumar. We also thank Selin Doganer for assistance in programming the psychophysics experiment and Alexander Lappe for helpful feedback on the manuscript.

\printbibliography

\end{document}